\documentclass{article}

\PassOptionsToPackage{numbers, compress}{natbib}

\usepackage[preprint]{arxiv}

\usepackage[utf8]{inputenc} 
\usepackage[T1]{fontenc}    
\usepackage{hyperref}       
\usepackage{url}            
\usepackage{booktabs}       
\usepackage{amsfonts}       
\usepackage{nicefrac}       
\usepackage{microtype}      
\usepackage{xcolor}         
\usepackage{amsmath}
\usepackage{amssymb}

\usepackage{amsthm}
\usepackage{commath}
\usepackage{comment}
\usepackage{tcolorbox}
\usepackage{subcaption}
\usepackage{algorithmic}
\usepackage{algorithm}

\theoremstyle{plain}
\newtheorem{theorem}{Theorem}[section]

\newtheorem{lemma}[theorem]{Lemma}

\theoremstyle{definition}
\newtheorem{definition}[theorem]{Definition}
\newtheorem{assumption}[theorem]{Assumption}
\theoremstyle{remark}

\title{On the Complexity of Learning to Cooperate with Populations of Socially Rational Agents}

\author{%
      Robert Loftin \\
      Department of Computer Science\\
      University of Sheffield \\
      Sheffield, S10 2TN, UK \\
      \texttt{r.loftin@sheffield.ac.uk} \\
      \And
      Saptarashmi Bandyopadhyay \\
      Department of Computer Science\\
      University of Maryland\\
      College Park, MD 20742, USA \\
      \texttt{saptab1@umd.edu} \\
      \And
      Mustafa Mert Çelikok \\
      Department of Intelligent Systems\\
      Delft University of Technology\\
      Delft, 2600 AA, The Netherlands \\
      \texttt{m.m.celikok@tudelft.nl} \\
      }

\begin{document}

\maketitle

\begin{abstract}

    Artificially intelligent agents deployed in the real-world will require the ability to reliably \textit{cooperate} with humans (as well as other, heterogeneous AI agents). To provide formal guarantees of successful cooperation, we must make some assumptions about how partner agents could plausibly behave.  Any realistic set of assumptions must account for the fact that other agents may be just as adaptable as our agent is.  In this work, we consider the problem of cooperating with a \textit{population} of agents in a finitely-repeated, two player general-sum matrix game with private utilities.  Two natural assumptions in such settings are that: 1) all agents in the population are individually rational learners, and 2) when any two members of the population are paired together, with high-probability they will achieve at least the same utility as they would under some Pareto efficient equilibrium strategy.  Our results first show that these assumptions alone are insufficient to ensure \textit{zero-shot} cooperation with members of the target population.  We therefore consider the problem of \textit{learning} a strategy for cooperating with such a population using prior observations its members interacting with one another.  We provide upper and lower bounds on the number of samples needed to learn an effective cooperation strategy.  Most importantly, we show that these bounds can be much stronger than those arising from a "naive'' reduction of the problem to one of imitation learning.
    
\end{abstract}

\section{Introduction}
\label{sec:introduction}

In this work, we address the problem of learning to cooperate with a \emph{socially intelligent} population of agents from observations interactions between members of this population.  We study cooperation in finitely-repeated, two-player, general-sum matrix games with private payoffs.  W say that a population of adaptive agents is socially intelligent if its members are (1) individually Hannan-consistent and (2) compatible in the sense that any pair of agents will perform nearly as well as some Pareto-optimal Nash equilibrium of the matrix game.  We argue that this model of cooperation is more realistic than those that assume identical payoffs or public utilities.  In real-world applications it is unlikely that independent agents will have identical utilities, or that they will provide complete information about their preferences or future behaviour to others.  In the case of AI--AI cooperation, agents developed by different companies will not have access to each other's source-code, while in the case of human--AI cooperation, having the human fully describe there preferences or behaviour in advance may be infeasible.  Therefore, the question we address in this work is: \emph{Can we learn to cooperate with a socially intelligent population of agent by observing its members cooperate with each other?} We answer this question by providing upper and lower bounds on the sample complexity of learning good cooperation strategies.

If we make no assumptions about the target population, we can do little more than attempt to mimic observed behavior as closely as possible, reducing the problem to one of imitation learning.  Unfortunately, the strategies of adaptive agents may depend on the full history of interaction, and so the sample complexity of imitation learning will grow exponentially in the length of the repeated game.  Our main contribution is an upper-bound showing that, for partners drawn from a socially intelligent (consistent and compatible) population, we can learn to cooperate with far fewer samples than would be required by a pure imitation learning approach.  

This result utilizes a class of what we refer to as \emph{imitate-then-commit} strategies, which leverage the fact that the population is socially intelligent to achieve cooperation without perfect imitation.  The key idea is that our agent only needs to learn to imitate a member of the target population long enough to for the average strategy to approximate a Pareto-efficient solution.  Once such a strategy is identified, our agent can switch to a \emph{coercive} strategy such that any Hannan-consistent partner will either continue to adhere to the current joint strategy, or else switch to a superior strategy, with either case corresponding to ``successful'' cooperation.

In section \ref{sec:preliminaries} formalize our repeated game setting, and provide background on external regret and Hannan-consistency.   We also propose a definition of cooperative compatibility (Definition~\ref{def:compatibility}) that is closely related to the notion of compatibility used in~\citep{powers2004targeted}.  In Section~\ref{sec:social_intelligence}, we provide our novel definition of social intelligence, and describe a realistic class of agents that satisfy it.  In Section~\ref{sec:learning} we formalize our learning problem as that of trying to minimize \emph{altruistic regret}, which we argue is the most natural measure of successful cooperation in this setting.  We also give lower bounds on its sample complexity under different sets of assumptions.  Finally, in Section~\ref{sec:upper_bound} we present an upper-bound on the number of samples needed to learn strategies that achieve small altruistic regret.


\section{Preliminaries}
\label{sec:preliminaries}

\paragraph{Repeated bi-matrix games with private types.} Let $i \in \{1,2\}$ denote the agent index. We assume both agents have $N$ pure strategies (henceforth "actions"). Let $\Theta$ denote the \textit{finite} type space, where $\theta_1, \theta_2 \in \Theta$ denote the \textit{private} types of the two agents, and $\boldsymbol{\theta} = (\theta_1, \theta_2)$ denotes the joint type.  We denote agent $i$'s payoff matrix as $G(\theta_i) \in \Re^{N \times N}$, and let $G(\boldsymbol{\theta}) = [G(\theta_1), G(\theta_2)^{\top}]$ denote the bi-matrix game parameterized by $\boldsymbol{\theta}$ (with agent 1 as the row player).  In a single \textit{episode}, the agents play $G(\boldsymbol{\theta})$ for a fixed number of stages $0 < T < \infty$.  We let $a^{1}_t$ and $a^{2}_t$ denote the actions chosen by agents 1 and 2 in stage $0 < t \leq T$.  For mixed strategies $\sigma, \sigma' \in \Delta(N)$, we let $G(\sigma, \sigma' ; \theta) = \sigma^{\top} G(\theta) \sigma'$.  We overload $a^{1}_t$ and $a^{2}_t$ to also denote the mixed strategies that assign all probability mass to actions $a^{1}_t$ and $a^{2}_t$, such that $G(a^{1}_t, a^{2}_t; \theta_1)$ and $G(a^{1}_t, a^{2}_t; \theta_2)$ are agent $1$ and $2$'s respective payoffs at stage $t$.  We also assume that for all $\theta \in \Theta$, $G_{ij}(\theta) \in [0, 1], \forall i,j \in [N]$. 

Let $\mathcal{H}_t = (N \times N)^t$ be the set of histories of length $t$ (with $\mathcal{H}_0 = \{ \emptyset \}$), and let $\mathcal{H}_{\leq t} = \bigcup^{t}_{s=0}\mathcal{H}_s$ be the set of all histories of length at most $t$.  The strategy space $\Pi$ for an agent is then the space of mappings $\pi : \Theta \times \mathcal{H}_{\leq T-1} \mapsto \Delta(N)$, where $\Delta(N)$ is the set of probability distributions over the action set $[N]$.  As a functional, a strategy $\pi$ maps each type $\theta$ to a \textit{behavioral strategy}~\cite[Chapter~5.2.2]{shoham2008multiagent} that maps histories of play to action distributions, such that $a^{i}_t \sim \pi_i (\theta_i, h_{t-1})$.  We denote agent $i$'s expected total payoff for following strategy $\pi$ against $\pi'$ as

\begin{equation}
    \label{eqn:expected_payoff}
    M_i (\pi, \pi' ; \theta, \theta') = \text{E}\left[ \left. \sum^{T}_{t=1} G(a^{i}_t, a^{-i}_t ; \theta_i) \right\vert \pi_i = \pi, \pi_{-i} = \pi', \theta_i=\theta, \theta_{-i}=\theta' \right],
\end{equation}
where the expectation is taken over the actions $a^{i}_t$ and $a^{-i}_t$ sampled from the agents' strategies.

\subsection{Consistency}
\label{sec:consistency}

A natural criterion for rationality is that an agent should attempt to to achieve a payoff nearly as large as the best response to its partner's average strategy, which we refer to as \textit{consistency}.  To account for the non-stationary behavior of other agents', we specifically consider \textit{Hannan consistency}~\cite{hannan1957consistency}, which in our finite-time setting simply requires that an agent have bounded \textit{external regret} over $T$ stages.  The external regret for agent $i$ is defined as
\begin{equation}
    \label{eqn:external_regret}
    R^{\text{ext}}_i (h ; \theta) = \max_{a^{i} \in [N]} \sum^{\vert h \vert}_{t=1} \left\{ G(a^{i}, a^{-i}_t(h) ; \theta_i) - G( a^{i}_t(h), a^{-i}_t (h) ; \theta_i) \right\}
\end{equation}
where $a^{i}_t(h)$ denotes the action $i$ played at stage $t$ within the history $h \in \mathcal{H}_{\leq T}$.

\begin{definition}[Consistency]
\label{def:consistency}
For $\delta, \epsilon, T > 0$, an agent $i \in \{1,2\}$ is $(\delta, \epsilon, T)$-\textit{consistent} if, for all types $\theta \in \Theta$, and \textit{any} partner strategy, we have that $\frac{1}{T} R^{\text{ext}}_i (h_T ; \theta) \leq \epsilon$ with probability at least $1-\delta$.
\end{definition}
\noindent We also define the \textit{expected} external regret $\bar{R}^{\text{ext}}_i (h ; \theta)$ by replacing the $a^{i}_t(h)$ (the action $i$ played at stage $t$) with their full strategy $\pi^{i}(\theta, h_t)$.  $R^{\text{ext}}_i (h ; \theta)$ and $\bar{R}^{\text{ext}}_i (h ; \theta)$ are related by the inequality
\begin{align}
    \label{eqn:azuma}
    R^{\text{ext}}_i (h_t ; \theta) &\leq \bar{R}^{\text{ext}}_i (h_t ; \theta) + \sqrt{\frac{T}{2} \ln\frac{1}{\delta}},
\end{align}
which holds w.p. at least $1-\delta$ for all $t \leq T$ simultaneously (this follows directly from~~\cite[Lemma~4.1]{cesa2006prediction}). We therefore only need to bound $\bar{R}^{\text{ext}}_i (h_t ; \theta)$ to provide high-probability regret bounds.

\subsection{Cooperative compatibility}
\label{sec:compatibility}

\begin{table}[t]
    \centering
    \begin{subtable}[t]{0.45\textwidth}
        \centering
        \begin{tabular}{c|c|c|}
                & $A$   & $B$ \\ \hline
            $A$ & $2,2$ & $0,0$ \\ \hline
            $B$ & $0,0$ & $1,1$ \\ \hline
        \end{tabular}
        \label{tab:games:coop}
        \caption{A fully-cooperative 2x2 matrix game.}
    \end{subtable}
    \begin{subtable}[t]{0.45\textwidth}
        \centering
        \begin{tabular}{c|c|c|}
                & $C$     & $D$ \\ \hline
            $C$ & $2,2$   & $0,3$ \\ \hline
            $D$ & $3,0$   & $1,1$ \\ \hline
        \end{tabular}
        \label{tab:games:prisoners}
        \caption{The prisoner's dilemma game.}
    \end{subtable}
    \caption{}
    \label{tab:games}
\end{table}

Even in a fully cooperative game, the fact that both agents are consistent does not guarantee that they will achieve an optimal outcome.  In the $2 \times 2$ game in Table 1a for example, both $(A,A)$ and $(B,B)$ are Nash equilibria to which consistent agents could converge, but only $(A,A)$ is optimal.  In general-sum games, consistency may preclude Pareto-optimal outcomes, as in the classic prisoner's dilemma game (Table 1b), where the only outcome in which neither player incurs positive regret is $(D,D)$, which is Pareto-dominated by $(C,C)$.Therefore, similar to~\cite{powers2004targeted}, we define successful cooperation in terms of the \textit{Pareto-optimal Nash equilibria} (PONE)~\cite{mas1995microeconomic} of a game $G$.  

Let $\mathcal{N}(G) \subseteq \Delta(N) \times \Delta(N)$ be the set of Nash equilibria (NE) of $G$.  For a fully-cooperative game, $\mathcal{N}(G)$ will contain all globally optimal strategy profiles for $G$.  It may, however, also contain joint strategies that are highly sub-optimal.  Let $\mathcal{P}(G) \subseteq \mathcal{N}(G)$ denote the set of Pareto optimal Nash equilibria.  In this work, we say that a strategy profile $\langle \sigma_1, \sigma_2 \rangle \in \mathcal{P}(G)$ if and only if $\langle \sigma_1, \sigma_2 \rangle  \in \mathcal{N}(G)$, and there does not exist $\langle \sigma'_1, \sigma'_2 \rangle  \in \mathcal{N}(G)$ such that $G(\sigma'_1, \sigma'_2; \theta_1) > G(\sigma_1, \sigma_2; \theta_1)$ \textit{and} $G(\sigma'_2, \sigma'_1; \theta_2) > G(\sigma_2, \sigma_1; \theta_2)$.  This means that $\langle \sigma_1, \sigma_2 \rangle$ is a PONE if it is a Nash equilibrium of $G$, and it is not \textit{strongly} Pareto-dominated by any other Nash equilibrium of $G$.  Intuitively, if two agents are individually consistent, and willing to cooperate with each other, their joint payoff profile should not be dominated by any PONE.  We formalize this intuition as follows:

\begin{definition}[Compatibility]
\label{def:compatibility}
    For $\delta, \epsilon, T > 0$, two agents $\pi^1$ and $\pi^2$ are $(\delta, \epsilon, T)$-\textit{compatible} if, when played together, for any joint type $\boldsymbol{\theta} \in \Theta \times \Theta$, w.p. at least $1-\delta$, $\exists \langle \sigma^{*}_1, \sigma^{*}_2 \rangle \in \mathcal{P}(G(\boldsymbol{\theta}))$ s.t.
   \begin{equation}
        \frac{1}{T}\sum^{T}_{t=1}  G(\sigma^{*}_{i}, \sigma^{*}_{-i} ; \theta_i) - G(a^{i}_t, a^{-i}_t ; \theta_i) \leq \epsilon,
  \end{equation}
  for both $i=1$ and $i=2$.
\end{definition}

A pair of agents is compatible if, when paired together, with high-probability over their path of play $h_T$ there will exist some PONE that does not $\epsilon$-dominate their realized payoffs.  Note that this definition is the approximate and finite-horizon version of the one provided in~\citep{powers2004targeted}.

\subsection{Socially intelligent agents}
\label{sec:social_intelligence}

We argue that it is natural to model an existing population of cooperating agents as a set of approximately compatible, but otherwise heterogeneous agents. We therefore introduce the more general idea of a socially intelligent \textit{class} of agents that are compatible with any other member of their class:

\begin{definition}[Social Intelligence]
\label{def:social_intelligence}
A set $C$ of agents forms a \textit{socially intelligent class} w.r.t. $\Theta$ if, for some $\delta, \epsilon, T > 0$, each agent $\pi \in C$ is $(\delta, \epsilon, T)$-consistent for all $\theta \in \Theta$, and any two agents $\pi, \pi' \in C$ are $(\delta, \epsilon, T)$-compatible over all joint types $\Theta$. An individual agent $\pi$ is called \textit{socially intelligent} if it forms a socially intelligent class $\{ \pi \}$ with itself.
\end{definition}

The Hannan consistency requirement ensures that any agent in the population always has bounded average regret, whereas the approximate compatibility means if both agents are from $C$, with high probability there will exist some PONE that does not $\epsilon$-dominate their path of play. Below we describe a socially intelligent class based on a pre-agreed \textit{coordination protocol}.

\paragraph{Coordination protocols} For a type space $\Theta$, we first define a function $s(\boldsymbol{\theta}) \in \mathcal{P}(G(\boldsymbol{\theta}))$ that maps from each joint type $\boldsymbol{\theta}$ to a strategy profile in $\mathcal{P}(G(\boldsymbol{\theta}))$.  We can think of $s(\boldsymbol{\theta})$ as a common ``convention'' the agents in $C$ have settled upon. Since we assume private types, members of $C$ do not know each other's type at the beginning of their interaction. If any type $\theta \in \Theta$ can be communicated to others in a sequence of $k < T$ actions, then agents in $C$ can agree on a coordination protocol similar to a handshake. Let the protocol be a map $\kappa(\theta)$ from types to a history-dependent policy. Then, at the beginning of each interaction, both agents will play $\kappa$ for $k$-steps in order to communicate their types.  After coordinating with each other, the agents play $s((\theta_i, \theta_{-i}))$ for the remaining $T-k$ steps.  The agents must still ensure their partner does not deviate from $s((\theta_i, \theta_{-i}))$ for safety against adversarial ``imposters''. Since playing a PONE jointly will lead to low regret for both, if $i$'s regret exceeds a certain threshold, this would indicate $-i$ is deviating from $s$ significantly. The threshold can be chosen by the aid of the following lemma,
\begin{lemma}
    \label{lem:nash}
    For any $\delta, T > 0$, if both players follow strategy $s(\boldsymbol{\theta})$ at each stage, then with probability at least $1 - \delta$ we have 
    \begin{equation}
        \bar{R}^{\text{ext}}_i (h_t ; \theta_i) \leq \sqrt{2T \ln \frac{2}{\delta}} \quad \text{and} \quad  R^{\text{ext}}_i (h_t ; \theta_i) \leq 2\sqrt{2T \ln \frac{4}{\delta}},
    \end{equation}
\end{lemma}
which follows from an application of the Azuma-Hoeffding inequality (shown in Appendix~\ref{apx:nash}). Then the question is what safe strategy should the $i$ fall back into, if the rule is triggered. We base the fallback strategy on the \textit{multiplicative weights}~\cite{freund1999adaptive} update rule, defined as:
\begin{equation}
    s^{i}_{\text{mw},k}(h_t ; \theta_i) \propto s^{i}_{\text{mw},k}(h_{t-1} ; \theta_i) \exp\left(-\eta G(k, a^{-i}_{t-1}(h); \theta_i) \right)
\end{equation}
for $k \in N$, where $s^{i}_{\text{mw}}(h_0 ; \theta_i)$ is the uniform strategy.  Define $\pi^{\text{mw},T}$ as the agent that plays $s^{i}_{\text{mw}}(h_t ; \theta_i)$ with learning rate $\eta = \sqrt{8 \ln(N / T)}$.  The expected external regret of $\pi^{\text{mw},T}$ is bounded as
\begin{equation}
    \bar{R}^{\text{ext}}_i (h_T ; \theta_i) \leq \sqrt{\frac{T}{2}\ln N}
\end{equation}
surely~\cite[Theorem 2.2]{cesa2006prediction}.  We then define the agent's overall strategy $\pi^{T,\epsilon}$ as follows:
\begin{enumerate}
    \item In first $k$ steps, play $\kappa(\theta_i).$ 
    \item If $-i$'s behaviour in $h_k$ not compatible with $\kappa(\theta)$ for any $\theta \in \Theta$, switch to $\pi^{\text{mw},T}$ for all subsequent stages.
    \item While $\bar{R}^{\text{ext}}_i (h_{t} ; \theta_i) \leq  k + \epsilon(T-k)  - \sqrt{\frac{T-k}{2}\ln N} - 1$, play $s_i(\boldsymbol{\theta})$.
    \item Otherwise, switch to $\pi^{\text{mw},T}$ for all subsequent stages.
\end{enumerate}
The theorem below shows that agents that follow the social authentication strategy above form a socially intelligent class among themselves. All proofs have been deferred to appendix \ref{apx:social_intelligence}.

\begin{theorem}
\label{thm:social_intelligence}
For any $\delta, T > k$, let $\epsilon_0 \geq \sqrt{\frac{2}{(T-k)} \ln\frac{2}{\delta}}$, and let $\epsilon_1 = \epsilon_0 + \sqrt{\frac{1}{2(T-k)} \ln N} + \frac{1}{(T-k)}$.  Then for $\epsilon = \epsilon_1 + \sqrt{\frac{(T-k)}{2}\ln\frac{1}{\delta}}$, the $\pi^{T,\epsilon_1}$ is $(\delta, \epsilon, T)$-socially intelligent.
\end{theorem}


\section{Learning to Cooperate}
\label{sec:learning}

Going forward, we will assume that our agent (henceforth referred to as the ``AI'') will take the role of agent $1$, while the other agent (referred to as the ``partner'') will be agent $2$.  Our goal is to choose a strategy for the AI that can cooperate with a partner drawn from some \textit{target population} nearly as effectively as agents from this population cooperate with one another.  For parametric game $G$, with type space $\Theta$, we will let the target population be a set $C$ of strategies forming an $(\delta,\epsilon,T)$-SI class w.r.t. $\Theta$.  Ideally, we would hope to choose an AI strategy $\pi$ that can cooperate with $C$ \textit{without} any additional information the strategies in $C$.  Looking at the coordination protocol example in Section~\ref{sec:social_intelligence}, we can see that in many cases a population is likely to use arbitrary conventions to coordinate their behavior, and intuitively we would imagine cooperation to be impossible without prior knowledge of these conventions. (We make this intuition formal in Theorem~\ref{thm:lower_bound}).

We therefore consider the problem of learning an cooperative AI strategy using prior observations of members of the target population interacting with one another.  We define a \textit{social learning problem} by a tuple $\{ G, \Theta, C, \rho, \mu \}$, where $C$ is the target population (SI w.r.t. $\Theta$), $\rho$ is a distribution over $C$, while $\mu$ is a distribution over the joint type space $\Theta \times \Theta$.  We can think of $C$ as the set of possible strategies that any member of the target population might follow, while $\rho$ is the frequency of those strategies within the population.  To choose an AI strategy, we leverage a dataset $\mathcal{D} = \{(\theta^j_1, \theta^j_2, h^j_T) | j \in [n] \}$ covering $n$ \textit{episodes} of length $T$.  In each episode $j$, two agents $\pi^1_j$ and $\pi^2_j$ are sampled independently from $\rho$, and played together under the joint type $\boldsymbol{\theta}_j \sim \mu$.  The AI observes the full history $h^j_T$, along with the agents' types $\theta^j_1$ and $\theta^j_2$.  We denote a specific learning algorithm as a data conditioned strategy $\pi(\mathcal{D})$.

\subsection{Altruistic Regret}
\label{sec:altruistic_regret}

We seek an AI strategy that minimizes the regret relative to some Pareto optimal solution to $G(\boldsymbol{\theta})$.  Rather than minimizing regret in terms of the AI's own payoffs, however, we seek to minimize \textit{partner's} relative to their (worst case) PONE in $G(\boldsymbol{\theta})$.  We formalize this regret with the following definition:

\begin{definition}[Altruistic Regret]
\label{def:altruistic_regret}
Let the $(\sigma^*_i, \sigma^*_{-i}) \in \mathcal{P}(G_{-i}(\theta_{-i}))$  denote the PONE with the \emph{lowest payoff} for the agent $-i$ where $i \in \{1,2\}$. The altruistic regret of agent $i$ is defined as     
 \begin{equation}
    R^{\text{alt}}_i(h_T; \theta_{-i})=\sum^{T}_{t=1}  G(\sigma^*_i, \sigma^*_{-i}; \theta_{-i}) - G(a^{i}(h_t), a^{-i}(h_t) ; \theta_{-i}).
  \end{equation}
\end{definition}

In practical cooperation tasks, we would expect outcomes that have low regret for the partner will have low regret for the AI as well.

The cooperation objective for the AI agent can then be formalized as minimising the altruistic regret. Unlike the definition suggests, the AI agent must know its own type as well. This is due to the fact that as seen in the coordination protocols example, if the AI fails to imitate a human of its type or fail to communicate its type correctly, the partner might switch to a safe strategy.

The goal for the AI is to minimize its \textit{expected} altruistic regret over partners sampled from $\rho$ and types sampled from $\mu$.  The following lemma shows that we can treat the problem of minimizing regret with respect to a heterogeneous population $C$ as that of minimizing regret w.r.t. a single stochastic strategy.

\begin{lemma}
    \label{lem:singleton}
    Let $C$ be a finite set of agents that are $(\delta, \epsilon, T)$-socially intelligent w.r.t. type space $\Theta$, and let $\rho$ be a distribution over $C$.  There exists a mixed strategy $\bar{\rho}$ that forms an $(\delta, \epsilon, T)$-socially intelligent class, and which is equivalent to playing against partners sampled from $\rho$ in expectation.
\end{lemma}
\emph{Proof.} In a perfect recall game, every behavioural strategy has an equivalent mixed strategy and vice-versa \cite{Aumann+1964+627+650}. Thus $\rho$ can equivalently be defined as a distribution over mixed strategies so that $\rho \in \Delta(\Delta(N))$. Then  defining $\Bar{\rho}(a) = \int_{\Delta(N)} \sigma(a) \, d\rho(\sigma)$ where $a \in [N]$ denotes a pure strategy (i.e. action) completes the proof.

In order to show the joint impact of consistency and compatibility on the learning problem, we discuss the cases where the population is either consistent or compatible, but not both. 

\subsection{Consistency without Compatibility}
\label{sec:consistency-without-compatibility}

Assume that $C$ consists of agents that are consistent but not necessarily compatible. The most general class in this case is the class of all no-external-regret learners (no-regret henceforth). 
It is a well-established result that the long-run average of no-regret learning converges to the set of coarse correlated equilibria. The question is whether the AI agent can learn to do better than a coarse correlated equilibrium when paired with a member of $C$, using only a dataset $\mathcal{D}$ that consists of histories of play for different CCEs.

\begin{theorem}
\label{thm:consistency}
There exists a consistent yet incompatible class of agents
$C$ such that even with an infinite amount of data, the AI cannot learn strategies that minimise altruistic regret.
\end{theorem}
\emph{Proof.} The proof follows from the theorem 3 of \citet{monnot2017limits} which shows that given any coarse correlated equilibrium of a two-player normal-form game, there exists a pair of no-regret learners that would converge to it. Since $C$ can be any subset of no-regret learners, we cannot exclude those who converge to inefficient CCE. If the class $C$ contains only the agents that converge to Pareto-inefficient CCE, we cannot hope to learn optimal strategies from any dataset. Given an observed CCE $z$ in the dataset, assume that the AI knows it is facing one of the two agents that generated $z$, but does not know their type explicitly. Using a Stackelberg argument similar to \citet{brown2024learning}, we prove in appendix \ref{apx:upper_bound} that the AI can compute and commit to a leader strategy such that the payoffs are never \emph{strongly} Pareto-dominated by $z.$ However even in this case, we cannot eliminate the possibility of it being weakly dominated.

Regardless of the dataset, in the online phase, the AI faces a new agent from $C$ each time and does not know their type. We may hope to learn a classifier to quickly infer our partner's type online from their behaviour, assuming there exists a mapping from initial behaviour to types. However, since $C$ consists only of no-regret learners guaranteed to converge to a CCE in self-play, they have no reason to initially communicate their types to each other.

\subsection{Compatibility without consistency}

Assume that the members of $C$ are compatible but not consistent. We can construct such a class by using the coordination protocols example from section \ref{sec:social_intelligence}. Now, when agents from $C$ successfully identify each other after the authentication phase, they proceed with playing the agreed-upon PONE. However, if at any moment they play the wrong action, there is no constraint on what strategy they will switch to. This setting is equivalent to the case considered by \citet{loftin2022impossibility} in their impossibility result. The members of $C$ can employ grim-trigger strategies that forever punish the other agent, triggered by a mistake at any point. Even if we eliminate grim-trigger strategies, the impossibility result has proven that there still exists strategies the members of $C$ can play once triggered, and make the other agent suffer regret arbitrarily close to $\frac{1}{2}$ with payoffs in $[0,1].$ Since a single mistake during the online interaction can lead to partner playing strategies that yield linear regret, the outsider must learn to imitate at least one member of $C$ perfectly from the dataset. Therefore the offline problem in this setting reduces to imitation learning, in particular the no-interaction case from \citet{rajaraman2020toward}.

For each agent, the authentication protocol $\kappa$ is equivalent to a history-dependent policy that they commit to playing in the first $k$ time-steps. The lower-bound on the expected sub-optimality of the imitation learning from \citet{rajaraman2020toward} is based on the fact that the imitator cannot do better than uniformly random in unseen states. In the case of $\kappa$, states correspond to histories up to length $k.$ Since every $k$-step history can be uniquely embedding a type, an unseen history means a high probability of making a mistake if paired with the corresponding type. Therefore, to avoid linear altruistic regret, the AI must observe at least $|\mathcal{H}_k|$ samples, where $\mathcal{H}_k$ is the set of all possible $k$-step histories. 
\begin{theorem}
Let $M$ be the number of unique samples of $k$-step histories in the dataset. There exists a class of agents $C$ with a $k$-step social authentication protocol such that to bound the probability of \emph{failing to authenticate}, we need $M \geq \frac{N^{3k} - \delta N^{3k} - N^{2k}}{N-1}$ samples. Then for growing $k$, the sample complexity lower bound is $M = \Omega(N^{2k}).$
\label{thm:sample-lower-imitate}
\end{theorem}
    \emph{Proof.} Consider the coordination protocol example mentioned above. Let $h_k \in \mathcal{H}_k$ be missing from the dataset. When the AI is paired with the corresponding partner type, the probability of correctly authenticating is $\frac{1}{N^{k}},$ and thus authentication fails with probability $\frac{N^k - 1}{N^k}.$ Assuming we face each type uniformly randomly, if we have $M$ unique samples, the probability of facing an unobserved history is $\frac{N^{2k} - M}{N^{2k}}$ since $|\mathcal{H}_k| = N^{2k}$. Then the probability of failing is $\frac{N^k - 1}{N^k} \times \frac{N^{2k} - M}{N^{2k}} = 1 - \frac{M}{N^{2k}} - \frac{1}{N} + \frac{M}{N^{3k}}.$ In order to bound this by $\delta,$ we need $M \geq \frac{N^{3k} - \delta N^{3k} - N^{2k}}{N-1}$ samples. Since $k$-steps need to embed each type uniquely, $k$ grows with the size of the type space. For large $k$, the bound is dominated by $N^{3k},$ thus we have $M = \Omega(N^{3k})$ as $k$ grows. 

An immediate conclusion that follows from theorem \ref{thm:sample-lower-imitate} is that for the case of compatibility without consistency, this sample complexity is for bounding the probability of suffering linear regret. This is due to the fact that failing to authenticate can now lead to linear regret, since the partner can switch to arbitrary strategies.

\subsection{Lower bound for socially intelligent populations}

\begin{theorem}
    \label{thm:lower_bound}
    Let $M$ denote the number of histories with unique first $k$-steps in dataset $\mathcal{D}$ generated by the members of a socially intelligent class $C$. There exists a $C$ where 
    $R^{\text{alt}}_i(h_T; \theta_{-i}) = T$ with
    probability $\frac{N^k - 1}{N^k} \times \frac{N^{2k} - M}{N^{2k}} = 1 - \frac{M}{N^{2k}} - \frac{1}{N} + \frac{M}{N^{3k}}.$
\end{theorem}
\emph{Proof:} Let $C$ be 
a socially intelligent class of agents following a coordination protocol akin to the one described in section \ref{sec:social_intelligence}. The probability follows from the proof of theorem \ref{thm:sample-lower-imitate} as the probability of failing to authenticate. If the authentication fails, the partner switches to an arbitrary Hannan-consistent strategy. As stated in section \ref{sec:consistency-without-compatibility}, a consistent partner strategy may never communicate the partner's type. Without knowing the partner's type, the agent's worst-case average altruistic regret can be $1$, since it cannot compute its true regret without the partner's type (see definition \ref{def:altruistic_regret}). Let there be two partner types $\theta_{-i} = \theta_2$ or $\theta_3.$ If the agent $i$ mistakenly assumes $\theta_{-i} = \theta_2$, its behaviour attempts to minimize $R^{\text{alt}}_i(h_T; \theta_{2}) = \sum^T_{t=1} G(\sigma^*_i, \sigma^*_{-i}; \theta_2) - G(a^{i}(h_t), a^{-i}(h_t) ; \theta_2)$. Meanwhile, the play of the partner will be a no-regret algorithm with respect to the external regret $R^{ext}_{-i}(h; \theta_3).$ Having no other constraints in the type space, there is nothing stopping us from constructing a $\Theta$ such that a strategy minimizing $R^{ext}_{-i}(h_T; \theta_3)$ ends up maximizing $R^{\text{alt}}_i(h_T; \theta_{2}).$ Imagine the ideal case of $R^{ext}_{-i}(h_T; \theta_3)=0$ where $-i$ plays the fixed best action in hindsight $a^*$ throughout $h_T.$ Then the altruistic regret observed by $i$ is $R^{\text{alt}}_i(h_T; \theta_{2}) = \sum^T_{t=1} G(\sigma^*_i, \sigma^*_{-i}; \theta_2) - G(a^{i}(h_t), a^{-i} = a^* ; \theta_2).$ Let $G(a^{i}, a^* ; \theta_2) = 0$ for all $a^{i}.$ Then the altruistic regret is $\sum^T_{t=1} G(\sigma^*_i, \sigma^*_{-i}; \theta_2)$ which is $T$ in the worst-case.


\section{Upper bound for socially intelligent populations}
\label{sec:upper_bound}

A key idea behind this work is that against a socially intelligent target population, rather than trying to imitate a member of the population perfectly throughout the entire episode, the AI only needs to imitate them long enough to learn about its partner's private type.  Once it has this information, the AI can leverage the fact that the partner's strategy is consistent against \textit{any} strategy, and try to ``coerce'' the human partner into playing a strategy that minimizes the altruistic regret.  We will refer to such strategies as \textit{imitate-then-commit} (IC) strategies, which use the previous observations $\mathcal{D}$ to learn an imitation strategy to follow over the first $\tilde{T} < T$ steps of the interaction.  In this section we provide an upper bound on the altruistic regret of a specific (IC) strategy, as a function of the number of episodes in $\mathcal{D}$, subject to the following assumptions:

\begin{assumption}
    \label{asm:social_intelligence}
    For $\delta, \epsilon > 0$, and some $\tilde{T} < T$, we have that
    \begin{enumerate}
        \item $\rho$ is $(\delta, \epsilon, T)$-consistent.
        \item $\rho$ is $(\delta, \epsilon, \tilde{T})$-compatible.
    \end{enumerate}
\end{assumption}


\paragraph{Imitation learning.}
Under an imitate-then-commit strategy, the sample complexity is defined entirely by the number of episodes the AI needs to observe to learn a good $\tilde{T}$-step imitation policy.  Fortunately, imitation learning is a well-studied problem, and we can largely leverage existing complexity bounds.  The one caveat is that in this setting we need bounds on the total variation distance between the distribution over the partial history $h_{\tilde{T}}$ under the population strategy $\rho$, and that under the learned strategy.  Given the dataset $\mathcal{D}$, we define the imitation strategy $\hat{\pi}^{1}_{\tilde{T}}(\mathcal{D})$ such that $\hat{\pi}^{1}_{\tilde{T}}(h; \theta, \mathcal{D})$ is the empirical distribution over agent $1$'s actions for each history-type pair $(h, \theta)$ occurring in $\mathcal{D}$, while $\hat{\pi}^{1}_{\tilde{T}}(h; \theta, \mathcal{D})$ is the uniform distribution over $N$ for $(h, \theta) \notin \mathcal{D}$.  We then define the \textit{marginal} strategy $\hat{\pi}^{1}_{\tilde{T}}$, which can be implemented by sampling a dataset $\mathcal{D}$, and then following the imitation strategy defined by $\mathcal{D}$ for the next $\tilde{T}$ steps.  We then have the following bound on the distribution of $h_{\tilde{T}}$ under the imitation strategy:

\begin{lemma}
    \label{lem:imitation_learning}
    Let $p_{\tilde{T}}$ be the distribution over partial histories $h_{\tilde{T}}$ under the population strategy $\rho$, and let $\hat{p}_{\tilde{T}}$ be their distribution under $\hat{\pi}^{1}_{\tilde{T}}$.  We have that
    \begin{equation}
        \Vert p_{\tilde{T}} - \hat{p}_{\tilde{T}} \Vert_{\text{TV}} \leq \min\left\{ \tilde{T}, \frac{N^{2(\tilde{T}+1)} \vert\Theta\vert \tilde{T}^2\log(K)}{K} \right\},
    \end{equation}
    where $K = \vert \mathcal{D} \vert$
\end{lemma}

This bound follows directly from that of~\citep{rajaraman2020toward} via Lemma 1 of~\citep{ciosek2022imitation} (see Appendix~\ref{apx:imitation_learning} for full proof).


\paragraph{Imitate-then-commit strategy.}
For history $h_{\tilde{T}} \in \mathcal{H}_{\tilde{T}}$, we let $\hat{z}(h_{\tilde{T}}) \in \Delta(N \times N)$ denote the empirical \textit{joint} strategy played up to and including step $\tilde{T}$.  We show that, using $\hat{z}(h_{\tilde{T}})$, it is possible to construct a \textit{mixture} $\nu$ over mixed strategies $x \in \Delta(N)$ that, in expectation over $\nu$, the partner's payoff under their best response to $x \sim \nu$ will be at least as large as their payoff under $\hat{z}(h_{\tilde{T}})$.  The corresponding IC strategy will operate as follows:
\begin{enumerate}
    \item Sample $\mathcal{D}$ and compute the imitation strategy $\hat{\pi}^{1}_{\tilde{T}}(\mathcal{D})$.
    \item Play $\hat{\pi}^{1}_{\tilde{T}}(\mathcal{D})$ for the first $\tilde{T}$ steps, and observe $h_{\tilde{T}}$.
    \item Compute a suitable mixture $\nu$ from $\hat{z}(h_{\tilde{T}})$, and sample $x \sim \nu$
    \item Sample actions from $x$ for the remaining $T - \tilde{T}$ steps
\end{enumerate}
We then have the following upper bound on the altruistic regret achievable with an imitate-then-commit strategy:

\begin{theorem}
    \label{thm:upper_bound}
    Given that Assumption~\ref{asm:social_intelligence} holds for $\rho$, there exists a data-dependent strategy $\pi^{\text{IC}}(\mathcal{D})$ such that when played by the AI as agent $2$, the altruistic regret satisfies
    \begin{equation}
        \label{eqn:upper_bound}
        \text{E}\left[ R^{\text{alt}}_{1}(h_T, \theta_{2}) \right] \leq 2\delta + \delta(K) + \left( 2\frac{T - \tilde{T}}{T} + 1\right) \epsilon,       
    \end{equation}
    where $K = \vert \mathcal{D} \vert$ and $\delta(K)$ is defined as
    \begin{equation}
        \delta(K) = \min\left\{ \tilde{T}, \frac{N^{2(\tilde{T}+1)} \vert\Theta\vert \tilde{T}^2\log(K)}{K} \right\}
    \end{equation}
    and where the expectation is taken over $h_T$, $\boldsymbol{\theta}$, and $\mathcal{D}$.
\end{theorem}

\paragraph{Proof sketch:} By Lemma~\ref{lem:imitation_learning}, we can learn an imitation strategy such that the corresponding distribution over $h_{\tilde{T}}$ and $\hat{z}(h_{\tilde{T}})$ is close to that under $\rho$ in self-play.  As $\rho$ is compatible, both agents' payoffs under $\hat{z}(h_{\tilde{T}})$ must be close to those under \textit{some} PONE.  Finally, we can construct a mixture $\nu$ for agent $1$ such that agent $2$'s payoffs under its (approximate) best-response are almost as large as those under $\hat{z}(h_{\tilde{T}})$ (see Appendix~\ref{apx:upper_bound}).


\section{Related Work}
\label{sec:related_work}

Our work is closely related to the previous targeted learning model~\cite{powers2004targeted, powers2005finite, chakraborty:icml10},which defines similar compatibility and consistency criteria. The notion of targeted optimality \cite{10.5555/3104322.3104348} include convergence to learning an approximately best response in a multi-agent model with high probability in a tractable number of steps against a population of memory-bounded adaptive agents. The main difference with our work is that targeted learning only requires consistency against a specific target class of partners, which generally would not include the agent itself, or other adaptive agents. We require socially intelligent agents to be consistent against all possible partner strategies. We also require that cooperation and consistent learning occur over a fixed time horizon $T$, rather than asymptotically.  These differences mean that a hypothetical ``universally cooperative'' agent might be able to leverage the consistency of its partner to achieve cooperation without a prearranged convention. Socially intelligent agents can modeled as individually rational learners \cite{loftin2023unifying} to achieve Pareto-efficient joint behavior. Our research builds on this work by considering a learning setting where the agent when paired with any member of the population will achieve at least the same utility with high probability as the Pareto-efficient approach.

The problem of training agents to be able to cooperate with previously unseen partners is sometimes referred to as \emph{ad hoc teamwork} \cite{stone2010ad, mirsky2022survey} or \emph{zero-shot coordination} \cite{hu2020other}, especially in the context of multiagent reinforcement learning. Many approaches in reinforcement learning train cooperative policies that are \textit{robust} to possible strategies that a human or an AI agent can follow \cite{carroll2019overcooked}.  A lot of these methods build a ``population'' of partner strategies and maximizes the diversity of this population in order to train the AI's policy against it \cite{strouse2021fictitious, cui2023adversarial}. Other approaches assume that there is no prior coordination between the agents \cite{hu2020other} to learn rational joint strategies while estimating the agents' mutual uncertainty about one-another's strategies \cite{treutlein2021lfc}. Ad-hoc multiagent coordination can be helpful to learn cooperation among AI agents with the ``other-play'' algorithm \cite{hu2020other} that  finds such a strategy as a solution to the corresponding \textit{label free coordination} problem \cite{treutlein2021lfc}. A possible approach to solve these problems can be self-play \cite{10.5555/3535850.3536105} where the agent can optimize themselves by playing with past iterations of themselves in order to estimate the strategies of unseen partners. However, the "self-play" approach can learn cooperative strategies which can "over-fit" \cite{NEURIPS2021_797134c3} to one another in the population of agents. A key goal of Ad hoc coordination (teamwork) and aligned research in zero-shot coordination work has been to avoid this type of overfitting \cite{NEURIPS2021_4547dff5}. Our problem domain is closely related to both \emph{ad hoc teamwork} or \emph{zero-shot coordination}, since we consider training an agent to cooperate with previously unseen partners, and assume no control over the partner. Even though population-based training approaches to ad hoc teamwork are common, they focus on fully cooperative environments such as Dec-POMDPs, where the main issue is creating a diverse enough population to train with \cite{rahman2024minimum}. We consider partners that are self-interested, and do not assume identical payoffs.

Finally, in the case of Hannan-consistent partners, our problem setting is closely related to strategizing against and learning to manipulate no-regret learners \cite{deng2019strategizing, brown2024learning}. This line of work studies whether an optimizer agent can achieve better payoff than CCE against no-regret learners by learning to enforce a Stackelberg equilibria on them. Their emphasis is on online learning and the optimizer's payoff, while we focus on the offline setting and cooperation.


\section{Conclusion}

We provide formal guarantees for successful and reliable cooperation of AI agents with populations of socially intelligent rational agents. This is based on the assumptions that 1) agents in the population are individually rational, and 2) agents in the population when cooperating with another agent in the same group can achieve, at least the same utility that they would with respect to some Pareto efficient equilibrium strategy. We formalize the notion of consistency and cooperative compatibility of agents in two-player general-sum finitely-repeated bi-matrix games between the agents and the population with private type. Our theoretical guarantees are in the offline cooperation setting where the agent has to cooperate with unseen partners in the population to strategize against and manipulate no-regret policies for which we formalize the idea of altruistic regret. We prove that the assumptions on its own are insufficient to learn \textit{zero-shot} cooperation with partners of the socially intelligent target population. We provide upper bounds on the sample complexity needed to learn a successful cooperation strategy along with lower bounds on when the multi-agent cooperation setting is needed with respect to the populations' trajectories, the state space and the length of the learning episodes. The bounds in these settings of the agent actively querying the MDP without knowing the transition dynamics of the population or the agent observing the populations' transition dynamics are much stronger than the bounds that can be derived by naively reducing the cooperation problem to one of reinforcement learning. These complexity analysis and formally proven bounds can be helpful to sustainably model the alignment problem of AI agents.

\bibliographystyle{unsrtnat}
\bibliography{arxiv_references}

\begin{ack}
This work has been supported by the Hybrid Intelligence Center, \url{https://hybrid-intelligence-centre.nl},
grant number 024.004.022. 
\end{ack}



\appendix


\section{Proofs for Section~\ref{sec:preliminaries}}
\label{apx:preliminaries}

\subsection{Proof of Lemma~\ref{lem:nash}}
\label{apx:nash}
Here the joint type $\boldsymbol{\theta}$ will be implicit.  For $i \in \{ 1, 2\}$, we define $V^{i}_t$ as
\begin{equation}
    V^{i}_t = G_i (s^{i}_t , s^{-i}_t) - G_i (s^{i}_t, a^{-i}_t)
\end{equation}
We can see that $\text{E}[V^{i}_t \vert h_{t-1}] = 0$.  We can then have that
\begin{align}
    \bar{R}^{\text{ext}}_t &= \max_{a \in N} \sum^{t}_{r=1} \left\{ G_i(a, s^{-i}_r) - G_i (s^{i}_r, a^{-i}_r) \right\} \\
    &= \max_{a \in N} \sum^{t}_{r=1} \left\{ G_i(a, s^{-i}_r) - G_i (s^{i}_r , s^{-i}_r) + G_i (s^{i}_r , s^{-i}_r) - G_i (s^{i}_r, a^{-i}_r) \right\} \\
    &= \sum^{t}_{r=1} \left\{ G_i (s^{i}_r , s^{-i}_r) - G_i (s^{i}_r, a^{-i}_r) \right\} = \sum^{t}_{r=1} V^{i}_r \\
    &\leq \sqrt{\frac{2}{T}\ln\frac{1}{\delta}}
\end{align}
with probability $1-\delta$ for all $t \leq T$ simultaneously.

This follows from the fact that $\vert V^{i}_t \vert \in [0,1]$ and the ``maximal'' Azuma-Hoeffding inequality~\cite{hoeffding1963probability}.  The second equality follows from the fact that $\langle s^{i}_t , s^{-i}_t \rangle = s(\theta)$ is a Nash equilibrium.  The first bound of Lemma~\ref{lem:nash} follows from a union bound over the probability for both players, while the second bound combines this with Equation~\ref{eqn:azuma}. $\square$

\subsection{Proof of Theorem~\ref{thm:social_intelligence}}
\label{apx:social_intelligence}

\begin{theorem}[2.6]
For any $\delta, T > k$, let $\epsilon_0 \geq \sqrt{\frac{2}{(T-k)} \ln\frac{2}{\delta}}$, and let $\epsilon_1 = \epsilon_0 + \sqrt{\frac{1}{2(T-k)} \ln N} + \frac{1}{(T-k)}$.  Then for $\epsilon = \epsilon_1 + \sqrt{\frac{(T-k)}{2}\ln\frac{1}{\delta}}$, the $\pi^{T,\epsilon_1}$ is $(\delta, \epsilon, T)$-socially intelligent.
\end{theorem}
\textit{Proof.} By the definition of $\epsilon_1$, $\pi^{T,\epsilon_1}$ will only deviate when playing with itself if at some point $ k < t \leq T$ one player incurs an expected external regret of at least $\epsilon_0$, and by Lemma~\ref{lem:nash} that will occur with probability at most $\delta$.  Therefore, $\pi^{T,\epsilon_1}$ is $(\delta, \epsilon_0, T)$-compatible.  We also have that the total expected external regret of the MW agent $\pi^{\text{mw},T}$ is at most $\sqrt{(T / 2) \ln N}$.  This means that if $\pi^{T,\epsilon_1}$ switches at stage $t$, then the maximum possible expected external regret incurred by $\pi^{T,\epsilon_1}$ will be less than $\bar{R}^{\text{ext}}_i (h_t ; \theta) + \sqrt{\frac{T}{2}\ln N}$. Since $\pi^{\text{mw},T}$ will always switch just before this point is reached, its total expected regret will be less than $\epsilon_1$ surely, and will be less than $\epsilon$ w.p. $1-\delta$.  As $\epsilon \geq \epsilon_0$, we have that the $\pi^{,T,\epsilon_1}$ is $(\delta, \epsilon, T)$-socially intelligent.











\section{Proofs for Section~\ref{sec:upper_bound}}
\label{apx:upper_bounds}

\subsection{Proof of Lemma~\ref{lem:imitation_learning}}
\label{apx:imitation_learning}

We first apply Theorem 4.4 of~\citep{rajaraman2020toward}, which states that, for episodic imitation learning over $H$-step trajectories, for any expert policy $\pi^*$ we have
\begin{equation}
    J(\pi^*) - \text{E}_{\mathcal{D}} \left[ J(\hat{\pi}^{1}_{\tilde{T}}(h; \theta, \mathcal{D})) \right] \leq \min\left\{ H, \frac{\vert S \vert H^2 \log(K)}{K}\right\},
\end{equation}
where $S$ is the state space, with per-step rewards bounded in $[0,1]$.  We can model the interaction with $\rho$ as a $\tilde{T}$-step episodic MDP/R with $S = \mathcal{H}_{\leq \tilde{T}}$.  Plugging in $H = \tilde{T}$, $\vert S \vert < N^{2(\tilde{T} + 1)}$, and $\pi^* = \rho$ gives us
\begin{equation}
    J(\rho) - \text{E}_{\mathcal{D}} \left[ J(\hat{\pi}^{1}_{\tilde{T}}(h; \theta, \mathcal{D})) \right] \leq \min\left\{ \tilde{T}, \frac{N^{2\tilde{T}} \vert\Theta\vert \tilde{T}^2\log(N)}{K} \right\}.
\end{equation}
This bound holds simultaneously for all possible reward functions bounded in $[0,1]$.  If we restrict the reward function $r$ to be non-zero only for the terminal states $\mathcal{H}_{\tilde{T}}$, we have
\begin{equation}
    J(\pi^*) - \text{E}_{\mathcal{D}} \left[ J(\hat{\pi}^{1}_{\tilde{T}}(h; \theta, \mathcal{D})) \right] = \text{E}_{p_{\tilde{T}}}[r(h_{\tilde{T}})] - \text{E}_{\hat{p}_{\tilde{T}}}[r(h_{\tilde{T}})],
\end{equation}
using the definition of the marginal strategy $\hat{\pi}^{1}_{\tilde{T}}$.  Finally, applying Lemma 1 of~\citep{ciosek2022imitation} gives us 
\begin{equation}
        \Vert p_{\tilde{T}} - \hat{p}_{\tilde{T}} \Vert_{\text{TV}} \leq \min\left\{ \tilde{T}, \frac{N^{2\tilde{T}} \vert\Theta\vert \tilde{T}^2\log(N)}{K} \right\},
\end{equation}
the desired result. \qedsymbol{}

\subsection{Proof of Theorem~\ref{thm:upper_bound}}
\label{apx:upper_bound}

First, let $\tau^{2}(\boldsymbol{\theta})$, defined as
\begin{equation}
    \tau^{2}(\boldsymbol{\theta}) = \min_{\langle \sigma^1, \sigma^2\rangle \in \mathcal{P}(G(\boldsymbol{\theta}))} G(\sigma^2, \sigma^1 ; \theta^2),
\end{equation}
denote agent $2$'s payoff under the worst possible payoff for a PONE of the game parameterized by joint type $\boldsymbol{\theta}$.  Let $\mathcal{C}$ denote the event that
\begin{equation}
     \tau^{2}(\boldsymbol{\theta}) - \frac{1}{\tilde{T}}\sum^{\tilde{T}}_{t=1}  G(a^{2}_t, a^{1}_t ; \theta_{2}) \leq \epsilon
\end{equation}
Because $\rho$ is $(\delta, \epsilon, \tilde{T})$-compatible, we have that $\text{Pr}_{\rho}\{ C \} \geq 1 - \delta$.  For $\delta(K)$ defined as
\begin{equation}
    \delta(K) = \min\left\{ \tilde{T}, \frac{N^{2(\tilde{T}+1)} \vert\Theta\vert \tilde{T}^2\log(K)}{K} \right\},
\end{equation}
Lemma~\ref{lem:imitation_learning} also gives us $\text{Pr}_{\hat{\pi}^1, \rho}\{ C \} \geq 1 - \delta - \delta(K)$.  We therefore have that
\begin{align}
    \text{E}_{\hat{\pi}^1, \rho}\left[ \sum^{T}_{t=1}  G(a^{2}_t, a^{1}_t ; \theta_{2}) \right] &\geq \text{E}_{\hat{\pi}^1, \rho}\left[ \sum^{T}_{t=1}  G(a^{2}_t, a^{1}_t ; \theta_{2}) \vert \mathcal{C} \right] - T(\delta + \delta(K)) \\
    = \text{E}_{\hat{\pi}^1, \rho}\left[ \sum^{\tilde{T}}_{t=1}  G(a^{2}_t, a^{1}_t ; \theta_{2}) \vert \mathcal{C} \right] &+ 
    \text{E}_{\hat{\pi}^1, \rho}\left[ \sum^{T}_{t=\tilde{T}+1}  G(a^{2}_t, a^{1}_t ; \theta_{2}) \vert \mathcal{C} \right] - T(\delta +\delta(K)) \\
    \geq  \text{E}_{\hat{\pi}^1, \rho}\left[ \sum^{T}_{t=\tilde{T}+1}  G(a^{2}_t, a^{1}_t ; \theta_{2}) \vert \mathcal{C} \right] &+ T(\tau^2(\boldsymbol{\theta}) - \epsilon - \delta - \delta(K))
\end{align}

We therefore need to lower-bound the term

\begin{equation}
    \text{E}_{\hat{\pi}^1, \rho}\left[ \sum^{T}_{t=\tilde{T}+1}  G(a^{2}_t, a^{1}_t ; \theta_{2}) \vert \mathcal{C} \right]
\end{equation}

This will be the expected payoff given the strategy $x \sim \nu$ the AI commits to for the remaining $T - \tilde{T}$ steps.  The idea now is that we can construct a mixture $\nu$ over strategies that the the AI can commit to for the remaining $T - \tilde{T}$ steps such that the partner's payoff under their (approximate) best-response will be nearly as good as that under $\hat{z}(h_{\tilde{T}})$.

Let $G(z ; \theta^2) = \sum_{i \in M} \sum_{j \in M} z_{i, j}G(j, i ; \theta^2)$ be agent 2's expected payoff under $z$.  For any joint strategy $z$, we can construct $\nu$ such that if the AI commits to strategies sampled from $\nu$, the partner will have the same information about the AI's probably actions as they would given their ``recommended'' action under $\hat{z}(h_{\tilde{T}})$.  We build on the construction used by~\citet{stengel2004leadership}.  For any joint strategy $z$, we let $z_j = \sum_{i \in N} z_{ij}$ denote the \textit{marginal} probability that the column player (agent 2) plays $j$ under $z$.  For all $j \in N$ such that $z_j > 0$, we define $x_j$ as the \textit{conditional} distribution over the row-player (agent 1's) actions given that the column player plays $j$, such that $x_j(i) = \frac{z_{ij}}{z_j}$.  We then define $\nu$ as the strategy that commits to each $x_j$ with probability $z_j$. 

We can show that, when the partner plays a best-response to $x \sim \nu$, their payoff will be no worse than under $z$ itself. We first construct a \textit{response function} $r_z$ such that when agent 2 responds to $x \sim \nu$ with $r_z(x)$, its expected payoff equals $G(z ; \theta^2)$.  Let $S = \{ j \in N : z_j > 0\}$, and partition $S$ into $\mathcal{P}$ such that, for each $P \in \mathcal{P}$, we have $x_j = x_l$ for all $j, l \in P$.  For each $P \in \mathcal{P}$, we then define the strategy $y_P$ such that

\begin{equation}
    y_P(j) = \frac{z_j}{\sum_{l \in P} z_l}  
\end{equation}

for each $j \in P$, with $y_P(j) = 0$ for $j \notin P$.  (Note that if $z$ corresponds to some \textit{uncorrelated} strategy $\langle x, y \rangle$, then $P = N$ and $y_P = y$.)  Finally, for $j \in S$, we define $P(j)$ as the partition containing $j$, and define $r_z$ such that $r_z(x_j) = x_{P(j)}$.  We leave $r_z$ undefined for $x$ where $\mu(x) = 0$.  Now let $x_P$ be the common conditional strategy for all $j \in P$, and let $z_P = \sum_{j \in P} z_j$.  We then have that

\begin{align}
    \text{E}_{x \sim \nu} G(r_z(x), x ; \theta^2) &= \sum_{j \in S} z_j \left[ x^{\top}_j G(\theta^2)^{\top} r_z(x_j) \right] \\
    &= \sum_{P \in \mathcal{P}}  z_P \left[ x^{\top}_P G(\theta^2)^{\top}  y_P \right] \\
    &= \sum_{P \in \mathcal{P}}  z_P \left( \sum_{i \in N} \sum_{j \in N} x^{\top}_P(i) y_P(j) G(\theta^2)^{\top}_{ij} \right) \\
    &= \sum_{P \in \mathcal{P}}  z_P \left( \sum_{i \in N} \sum_{j \in N} \text{Pr}_z \{ i \vert P \} \text{Pr}_z \{ j \vert P \} G(\theta^2)^{\top}_{ij} \right)\\
    &= \sum_{P \in \mathcal{P}}  z_P \left( \sum_{i \in N} \sum_{j \in N} \text{Pr}_z \{ i, j \vert P \} G(\theta^2)^{\top}_{ij} \right) \\
    &= \sum_{i \in N} \sum_{j \in N} z_{ij} G(\theta^2)^{\top}_{ij} = G(z ; \theta^2)
\end{align}

where we have used the fact that $i$ and $j$ are independent given that $j \in P$.  Next, we have that for any best-response function $r^*$, we have

\begin{equation}
    \begin{split}
    G(z ; \theta^2) & = \text{E}_{x \sim \nu} G(r_z(x), x ; \theta^2) \\
                    & = \text{E}_{x \sim \mu}[ x^{\top} G(\theta^2)^{\top} r_z(x) ] \\
                    & \leq \text{E}_{x \sim \mu}[ \max_{y \in \Delta(N)} x^{\top} G(\theta^2)^{\top} y ] \\
                    & = \text{E}_{x \sim \mu}[ x^{\top} G(\theta^2)^{\top} r^*(x) ] \\
                    & = \text{E}_{x \sim \nu} G(r^*(x), x ; \theta^2)
    \end{split}
\end{equation}

Therefore, so long as the partner plays a best-response to the AIs chosen strategy, the will achieve at least the same payoff (in expectation) as they would under the strategy $z$ from which $\nu$ was computed.  Note however that $\rho$ will be (approximately) consistent over the full $T$ steps, not just the last $T - \tilde{T}$.  Define $\alpha = \frac{\tilde{T}}{T}$ and $\beta = \frac{T - \tilde{T}}{T}$, and let $z^1$ be agent 1's marginal strategy under $z$.  With probability $1-\delta$, $\rho$ will play an $\epsilon$-best-response to the mixture $\alpha \hat{z}(h_{\tilde{T}})^1 - \beta x$, with $x \sim \nu$.

Let $\mathcal{C}'$ be the event that $\rho$ is $\epsilon$-consistent over $T$ steps.  We then have that
\begin{align}
    \text{E}_{\hat{\pi}^1, \rho}\left[ \sum^{T}_{t=\tilde{T}+1}  G(a^{2}_t, a^{1}_t ; \theta_{2}) \vert \mathcal{C} \right] &\geq \text{E}_{\hat{\pi}^1, \rho}\left[ \sum^{T}_{t=\tilde{T}+1}  G(a^{2}_t, a^{1}_t ; \theta_{2}) \vert \mathcal{C}, \mathcal{C}' \right] - T\delta \\
    &\geq (T - \tilde{T})\left( \tau^{2}(\boldsymbol{\theta}) - 2\epsilon\right) - T\delta
\end{align}

Finally, dividing by $T$ and subtracting from $\tau^2(\boldsymbol{\theta})$, we get
\begin{equation}
    \text{E}\left[ R^{\text{alt}_{1}}(h_T, \theta_{2}) \right] \leq 2\delta + \delta(K) + \left( 2\frac{T - \tilde{T}}{T} + 1\right) \epsilon
\end{equation}
the desired result.

\qedsymbol{}

\end{document}